\documentclass[runningheads]{llncs}

\usepackage{lipsum}

\newcommand\blfootnote[1]{%
  \begingroup
  \renewcommand\thefootnote{}\footnote{#1}%
  \addtocounter{footnote}{-1}%
  \endgroup
}

\usepackage[utf8]{inputenc}
\usepackage[english]{babel}
\usepackage{color}
\usepackage{graphicx}
\usepackage{wrapfig}
\usepackage{wrapfig,lipsum,booktabs}
\usepackage[fleqn]{amsmath}

\begin{document}
\title{Deep Chronnectome Learning via Full Bidirectional Long Short-Term Memory Networks for MCI Diagnosis}
\titlerunning{Deep Chronnectome Learning via Full-BiLSTM for MCI Diagnosis}
%
\author{Weizheng Yan\inst{1,2,3} \and
Han Zhang\inst{3} \and
Jing Sui\inst{1,2}\and
Dinggang Shen\inst{3}}
\authorrunning{W. Yan \& H. Zhang et al.}
%
\institute{Brainnetome Center and National Laboratory of Pattern Recognition, 
Institute of Automation, Chinese Academy of Sciences, China \and
University of Chinese Academy of Sciences, China \and
Department of Radiology and BRIC, University of North Carolina of Chapel Hill, Chapel Hill, NC, USA\\
\email{dgshen@med.unc.edu}}
\maketitle       
\begin{abstract} \blfootnote{W. Yan and H. Zhang contribute equally to this paper} 
Brain functional connectivity (FC) extracted from resting-state fMRI (RS-fMRI) has become a popular approach for disease diagnosis, where discriminating subjects with mild cognitive impairment (MCI) from normal controls (NC) is still one of the most challenging problems. Dynamic functional connectivity (dFC), consisting of time-varying spatiotemporal dynamics, may characterize ``chronnectome'' diagnostic information for improving MCI classification. However, most of the current dFC studies are based on detecting discrete major ``brain status'' via spatial clustering, which ignores rich spatiotemporal dynamics contained in such chronnectome. We propose \textit{Deep Chronnectome Learning} for exhaustively mining the comprehensive information, especially the hidden higher-level features, i.e., the dFC time series that may add critical diagnostic power for MCI classification. To this end, we devise a new Fully-connected \textit{bidirectional} Long Short-Term Memory (LSTM) network (Full-BiLSTM) to effectively learn the periodic brain status changes using both past and future information for each brief time segment and then fuse them to form the final output. We have applied our method to a rigorously built large-scale multi-site database (i.e., with 164 data from NCs and 330 from MCIs, which can be further augmented by 25 folds). Our method outperforms other state-of-the-art approaches with an accuracy of 73.6\% under solid cross-validations. We also made extensive comparisons among multiple variants of LSTM models. The results suggest high feasibility of our method with promising value also for other brain disorder diagnoses.

\end{abstract}

\section{Introduction}
\vspace{-0.2cm}  
Alzheimer’s Disease (AD) is an irreversible neurodegenerative disease leading to progressive cognitive and memory deficits. Early diagnosis of its preclinical stage, mild cognitive impairment (MCI), is of critical value as timely treatment could be the most effective during this stage. Resting-state functional MRI (RS-fMRI) provides an opportunity to assess brain function non-invasively and has been successfully exploited to identify MCI~\cite{ref3}.To capture the time-varying information brain networks, dynamic functional connectivity (dFC) was proposed to characterize the time-resolved connectome, i.e., chronnectome, mostly using sliding-window correlation approach~\cite{ref4,ref6}. While promising, many current studies have not deeply exploited the rich spatiotemporal information of the chronnectome and utilized it in classification. For example, many studies focused on group comparison by detecting a set of discrete major brain status via clustering time-resolved FC matrices and further calculating their occurrence and dwelling time~\cite{ref6}. Inspired by the new finding that the brain dynamics are hierarchically organized in time(i.e., certain networks are more likely to occur preceding and/or following others~\cite{ref8}), we propose to learn diagnostic features in an end-to-end deep learning framework to better classify MCI.

Recurrent neural networks (RNNs) is a powerful neural sequence learning model for time series analysis.  LSTMs are improved RNNs that can effectively solve the “gradient exploding/vanishing” problem by controlling information flow with several gates~\cite{ref10}. It has recently been demonstrated to be able to handle large-scale learning in speech recognition and language translation tasks~\cite{ref11}. However, there is still a significant gap between brain chronnectome modeling and common time series analysis. Directly applying LSTM to dFC-based MCI diagnosis is non-trivial:\textit{1)} Brain is extraordinary complex whose dynamics could be substantially different from natural language interpretation.\textit{2)} The background noise is usually more intense in the brain dFC signals, compared to audio/video signals, making it very difficult to capture.\textit{3)} The brain may continuously use contextual information for guiding higher-level cognitive functions rather than produce an output at the end of the time series with a strict direction. Therefore, a general LSTM could not be suitable for brain chronnectome-based classification. To solve this problem, we propose a new deep learning framework that changes the traditional LSTM in two aspects. \textit{First}, we create Full-LSTM that connects the outputs of all cells to a “fusion” layer to capture a common time-invariant status-switching pattern, based on which the MCI can be diagnosed. \textit{Second}, to excavate the contextual information hidden in the dFC, we further use a bidirectional LSTM (BiLSTM) to access long-range context in both directions~\cite{ref12}. We hereby come out with an end-to-end chronnectome-based classification model, namely \textit{Full-BiLSTM}. The performance of our proposed method has been compared with state-of-the-art methods on ADNI-2 database. As the first “Deep Chronnectome Learning” study, we comprehensively compared the performance of three variants of LSTMs and reported the effect of different hyperparameters. The results support our hypothesis and significantly improved MCI diagnosis.
\vspace{-0.2cm}  

\begin{wrapfigure}{l}{0.69\textwidth}
    \centering
    \includegraphics[width=8cm, height=6cm]{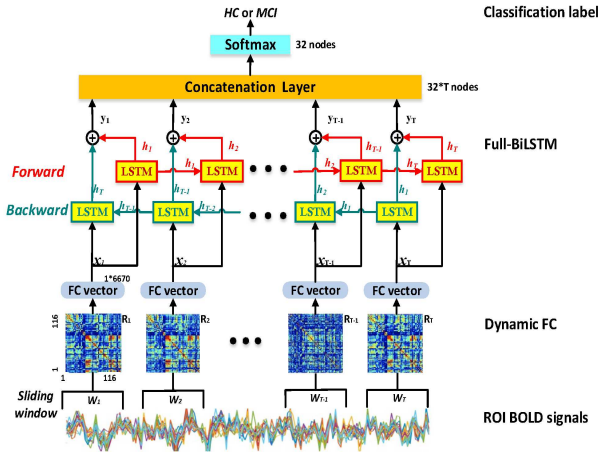}
    \caption{Overview of the Full-BiLSTM for MCI classification.}
     \label{Fig1}
    \vspace{-0.6cm}  
\end{wrapfigure}

\section{Methods}

First, the dFC of each subject is calculated using a sliding window-based method. Second, the dFC vector is used as the input of the BiLSTM module. Finally, the outputs of each repeating BiLSTM cell are concatenated into a dense layer for further prediction.

\subsection{Computing  dFC via a Sliding Window Method}

For each subject, the whole-brain time-varying connectivity matrices are computed based on ${M(M=116)}$ ROIs from the automated anatomical labeling (AAL) template using a sliding window approach~\cite{ref5,ref6}. As shown in Fig.~\ref{Fig1}, the averaged BOLD time-series ${S_i}$ in ROI $i$ are first computed. Then, the window ${\{W_t\} }$  are generated and applied to ${S=\{S_i\}}$, where ${T}$ is the total number of sliding windows. Next, for each ${W_t}$, an FC matrix ${R_t}$ of size ${M*M}$ that includes FC strengths between all pairs of ${S_{it}}$  are calculated. Thus, for each subject, a set of ${R_t (t=1,2,…,T)}$ are obtained, representing the subjects' whole-brain dFC. Due to the symmetry of each ${R_t}$, all FC strengths in ${R_t}$ among ${M}$ ROIs corresponding to a window ${t}$ are converted to a vector ${x_t}$ with ${M(M-1)/2}$ elements. Therefore, all the dFC time series from the ${k_{th}}$ subject can be represented by a matrix $X^k=[x_1^k,x_2^k,…,x_t^k]$ with a size of $T*\{M(M-1)/2\}$ and used as input to Full-BiLSTM classification model.

\subsection{Fully-Connected Bidirectional LSTM (Full-BiLSTM) }

\subsubsection{Long Short-Term Memory (LSTM).} LSTMs incorporates recurrently connected units, each of which receives an input $h_{t-1}$  from its previous unit as well as the current input $x_t$ for the current time point t. Each unit has its memory updating the previous memory $c_{t-1}$ with the current input modulation. The network takes three inputs: $x_t$,$h_{t-1}$,and $c_{t-1}$, and has two outputs: $h_t$ (the output of the current cell state) and $c_t$ (the current cell state). Three gates separately controls input, forget, output. The unit can be expressed as:
    \vspace{-0.2cm}
    \begin{equation} \label{eq1} 
        Input \ Gate: i_t= \sigma{(W_{xi} x_t+W_{hi} h_{t-1}+b_i )} 
    \end{equation} 
    \begin{equation}\label{eq2} 
        Forget \ Gate: f_t= \sigma{(W_{xf} x_t+W_{hf} h_{t-1}+b_f )} 
    \end{equation}
    \begin{equation}\label{eq3}
	    Output\ Gate: o_t=\sigma{(W_{xo} x_t+W_{ho} h_{t-1}+b_o)}
    \end{equation}
    \begin{equation}\label{eq4}
	    Input\ Modulation:g_t= \phi{(W_{xc} x_t+W_{hc} h_{t-1}+b_c)}  
    \end{equation}
    \begin{equation}\label{eq5}
	    Memory\ Cell\ Update: c_t=i_t\odot g_t+f_t \odot c_{t-1} 
    \end{equation}
    \begin{equation}\label{eq6}
    	Output: h_t = o_t \odot \tanh(c_t)
    \end{equation} 
Specifically, the input gate $i_t$ controls how much influence the inputs $x_t$ and $h_{t-1}$ exerts to the current memory cell (Eq.\ref{eq1}). The forget gate $f_t$ controls how much influence the previous memory cell $c_{t-1}$ exerts to the current memory cell $c_t$ (Eq.\ref{eq2}). Output gate controls how much influence the current cell $c_t$ has on the hidden state cell $h_t$ (Eq.\ref{eq3}). The memory cell unit $c_t$ is a summation of two components: the previous memory cell unit $c_{t-1}$, which is modulated by $f_t$ and $g_t$ (Eq.\ref{eq4}), and a weighted combination of the current input and the previous hidden state, modulated by the input gate $i_t$ (Eq.\ref{eq5}). Likewise, cell state is filtered with the output gate ${o(t)}$ for a hidden state updating (Eq.\ref{eq6}), which is the final output from an LSTM cell. With the inputting dFC time series, $W_{x\cdot}$ matrices (containing weights applied to the current input) and ${W_{h\cdot}}$ matrices (representing weights applied to the previous hidden state) can be learned, ${b_{\cdot}}$ vectors are biases for each layer, $\sigma$ is sigmoid, $\phi$ is $\tanh$ function, and$\odot$ denotes element-wise multiplication.

\vspace{-0.5cm} 
\subsubsection{Bidirectional LSTM (BiLSTM).}
BiLSTM is an effective solution that gets access to both preceding and succeeding information (i.e., context) by involving two separate hidden layers with opposite information flow directions~\cite{ref13}. For a brief description, we denote a process of an LSTM cell as ${H}$. BiLSTM first computes the forward hidden $\overrightarrow{h}$ and the backward hidden sequence $\overleftarrow{h}$ separately(Eq.\ref{eq7}-\ref{eq8}), and then combines $\overrightarrow{h_t}$ and $\overleftarrow{h_t}$ to generate the final output ${y_t}$ (Eq.\ref{eq9}). The ${W_{x\cdot}}$ and ${W_{h\cdot}}$ matrices in (Eq.\ref{eq7}-\ref{eq8}) are the same as those in(Eq.\ref{eq1}-\ref{eq4}). The ${W_{\overrightarrow{h}y}}$  (representing weights applied to the forward hidden state) and ${W_{\overleftarrow{h}y}}$ (representing weights applied to the backward hidden state) are learned with the inputting dFC time series. ${b_{\cdot}}$ vectors are biases for each layer.
\begin{equation}\label{eq7}
Forward\ LSTM: \overrightarrow{h}_t=H(W_{x\overrightarrow{h}}x_t + W_{\overrightarrow{hh}\overrightarrow{h}_{t-1}} + b_{\overrightarrow{h}})
\end{equation}
\vspace{-0.3cm}
\begin{equation}\label{eq8}
Backward\ LSTM: \overleftarrow{h}_t=H(W_{x\overleftarrow{h}}x_t + W_{\overleftarrow{hh}\overleftarrow{h}_{t-1}} + b_{\overleftarrow{h}}) 
\end{equation}
\begin{equation}\label{eq9}
Combined\ Output: y_t = H(W_{\overrightarrow{h}y}\overrightarrow{h}_t + W_{\overleftarrow{h}y\overleftarrow{h}_{t}} +b_y)  
\end{equation}
\vspace{-1cm}

\subsubsection{Full-BiLSTM.}
The traditional BiLSTM classification model usually uses the final state ${y_T}$ for classification~\cite{ref12}. However, this is insufficient for chronnectome-based diagnosis, because brain may continuously use contextual information to facilitate higher-level cognition and guide status transition, rather than producing a single output at the end of the scanning period. Therefore, the outputs of every repeating cell could be of equally important use and should be concatenated into a dense layer ${Y=[y_1,…y_t,{\ldots},y_T ]}$ (see “Concatenation Layer” in Fig.~\ref{Fig1}).). With this layer, we may abstract a common and time-invariant dynamic transition pattern from all the BiLSTM cells which may represent a constant ``trait'' information of each subject, instead of the continuously varying brief brain status. While the latter could be of great use in previous status-based studies such as those used \textit{Hidden Markov Chain} for status transition probability modeling in group-level comparison studies~\cite{ref8}, it will inevitably lose the precious temporal information which could capture more subtle individual differences for the more challenging disease diagnosis studies. In our framework for MCI diagnosis, the dense layer ${Y}$ is followed with softmax layer to get the final classification result. 

\vspace{-0.2cm}  

\section{Experiments and Results}
\subsection{Data Preprocessing}

In this study, we use the publicly available Alzheimer’s Disease Neuroimaging Initiative dataset (ADNI) to test our method. As shown in Table 1, 143 age- and gender-matched subjects (48 NCs with 164 RS-fMRI scans, and 95 MCIs with 330 RS-fMRI scans) were selected from ADNI-2 database. The goal of ADNI-2 study is to validate the use of various biomarkers including RS-MRI to find the best way to diagnose AD at pre-dementia stage. Each RS-fMRI scan was acquired using 3.0T Philips scanners at different medical centers. All the data were carefully reviewed by the quality control team in Mayo Clinic. ADNI is to date the largest, multi-site, rigorously controlled early AD diagnosis data. The RS-fMRI data were preprocessed following the standard procedure~\cite{ref3}.

\vspace{-0.2cm}  
\subsection{Dynamic Functional Connectivity Matrix}
In this experiment, the window length was 90s (30 volumes) as suggested by previous dFC studies ~\cite{ref6}. The window slides in a step of 2 volumes (6s), resulting in 54 segments of BOLD signals. For each subject and each scan, 54 FC matrices were obtained, reflecting the chronnectome. The upper half of the matrix containing 6670 unique dFC links were used and then reshaped into $X^k$ with the size of $54*6670$.

\begin{wraptable}{r}{6.5cm}
\vspace{-1.5cm}
\caption{Demographic information.}\label{tab1}
\begin{tabular}{lll}
\hline
 &  NC & MCI\\
\hline
Number of scans &  164 & 330\\
Age(mean($ \pm $ std,yrs)) &  75.4$  \pm $ 6.2 & 72.0 $ \pm $ 7.5\\
Gender(M/F) & 72/92 & 178/152\\
\hline
\vspace{-1cm} 
\end{tabular}
\end{wraptable}

\vspace{-0.2cm}  
\subsection{Data Augmentation}

Training deep learning models requires a large number of samples. Fortunately, only part of the dFC time series might be sufficient for discriminating MCIs from NCs because the FC dynamics could happen in a very brief period~\cite{ref8}. This allows us to conduct data augmentation to increase the sample size. Specifically, for each $X^k$, a continuous submatrix of length 30 were cropped as a new sample. By using a sliding window strategy with a stride of 1, the original $X^k$ can be augmented for ${54-30+1=25}$ times (augmented by a factor of 25). The label of the augmented data from the same subject was kept the same. Of note, all augmented sequences belonging to the same subject were used solely in the training, or validation, or testing phase. In the testing phase, the predicted labels for all the augmented data from the same subject was derived with majority voting to determine the final label for this subject.

\vspace{-0.2cm}  
\subsection{Full-BLSTM Parameters and Training Strategy} 

The Full-BiLSTM model was trained and evaluated using Keras. Data was split into 80\% for training and 20\% for testing (5-fold cross-validation). 10\% of samples from training data were further selected for validation to monitor the training procedure. Training was stopped when the validation loss stopped decreasing for 20 epochs or when the maximum epochs had been executed. The testing data was applied to the trained model to evaluate the performance. The model was trained for minimizing the weighted cross-entropy loss function using stochastic gradient descent (SGD) optimizer. The learning rate (lr) was started from 0.001 and decayed over each update as follow: ${lr_t=lr_{t-1}/(1+decay_{rate}*epochs)}$. The $decay_{rate}$ was $10^{-6}$, and the maximum epochs was 200. The batch size was 32. The weights and biases were initialized randomly. To improve the generalization performance of the model and overcome the overfitting problem, we used a dropout method ${(dropout=0.5)}$ and $l_1 norm$ regularization ${(l1=0.0005)}$. 

\vspace{-0.2cm}  
\subsection{Method Comparison}
As dFC is novel in this field, the disease diagnosis works using dFC are quite limited. We compared our approach against various classifiers commonly used. The majority of the dFC studies focus on brain statuses detected by clustering, or the temporal variability of dFC series. Therefore, in the competing methods, we also use these two types of the dFC features for MCI classification. In summary, we compared our method with the classification models using: 1) static FC (sFC); 2) dFC-based brain statuses~\cite{ref6}; and 3) dFC variability~\cite{ref3}, as detailed below. 
\vspace{-0.5cm}  

\subsubsection{sFC.}
The traditional FC method used in most of the FC studies are based on Pearson’s correlation of full-length BOLD signals. After building sFC matrix, an SVM classifier is trained based on the sFC strengths.
\vspace{-0.5cm}  

\subsubsection{Status-based.}
Group-level chronnectome status is identified by using k-means clustering with all of the dFC matrices in the training data. The occurrence frequency of each status is computed to as features. Then, an SVM classifier is constructed based on the frequency features of all status.
\vspace{-0.5cm}  

\subsubsection{Variability-based.}
Based on the dFC matrices, the quadratic mean value is computed for each dFC. A total of 6670 features are generated for each subject representing the fluctuation of the signals. The features are further reduced using two-sample t-test. An SVM classifier is constructed based on the dFC variability features.
\begin{table}
\centering
\vspace{-0.2cm}
\caption{Performance of different methods in MCI/NC classification.}\label{tab2}
\vspace{-0.3cm}
\begin{tabular}{|l|l|l|l|l|l|l|}
\hline
Method & ACC(std)\% & SEN(std)\% & SPE(std)\% & f1(std)\% & AUC(std)\% \\ 
\hline
\textcolor{blue}{Static FC + SVM} &  61.5(10.0) &	74.0(9.2) &	41.7(14.0) & 70.9(8.2) & 64.2(10.8)\\
\textcolor{blue}{dFC-variability}	& 54.8(12.9) &	54.4(12.3) & 56.8(19.1) &60.5(12.3)&	49.0(17.0)\\
\textcolor{blue}{dFC-status} &	61.3(10.0)&	70.8(12.2)&	47.2(13.6)&	69.9(8.6)&	61.9(15.9)\\
\textcolor{red}{\textit{Full-LSTM32}}	 &71.9(5.9)&	72.3(7.9)&	70.5(15.1)&	76.2(5.3)&	75.9(5.8)\\
\textit{Full-BiLSTM32-Stack}	&69.0(5.0)&	66.7(4.7)&	73.0(9.2)&	73.1(3.5)&	79.2(2.7)\\
\textcolor{red}{\textit{BiLSTM32-Last}} &	71.0(10.3)&	76.8(9.6)	&60.9(12.8)&	76.7(8.8)&	75.9(6.0)\\
\textbf{\textit{Full-BiLSTM32}} &	73.6(3.7)&	73.9(10.1)&	73.5(7.3)&	77.6(4.4)&	79.8(6.9)\\
\hline
\end{tabular}

\vspace{-0.2cm}
\begin{flushleft}
{\scriptsize Notes: Blue-colored methods are the traditional methods; Methods in italic are LSTM-based methods; Our method is in bold italic; Red italic indicates the model without bi-directional LSTM or without Full-LSTM.}
\end{flushleft}

\end{table}

\begin{figure}
\centering %
\vspace{-0.5cm}
\begin{minipage}[b]{0.5\textwidth} %
\centering %
\includegraphics[width=0.8\textwidth]{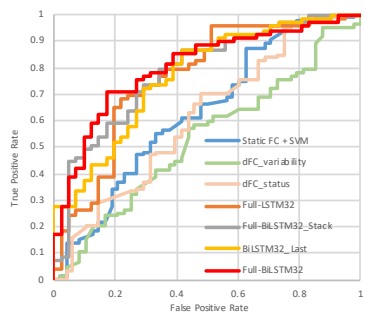} 
\caption{ROC curves of different methods.}
\label{Fig3}
\vspace{-0.5cm}
\end{minipage}
\begin{minipage}[b]{0.49\textwidth} %
\centering %
\includegraphics[width=0.8\textwidth]{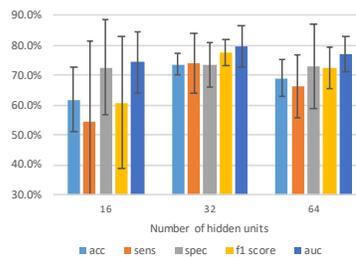}%
\caption{Effect of different hidden units}
\label{Fig4}
\vspace{-0.5cm}

\end{minipage}
\end{figure}

The performance comparison results are summarized in Table~\ref{tab2} and Fig.~\ref{Fig3} showing the ROI curves of all methods. Because of sample imbalance, the area under the ROC curve (AUC) was used as the main metric for comparing the performance of all the methods. Our method achieved 79.8\% in AUC and significantly outperformed the traditional sFC and dFC methods. The dFC variability method achieved the lowest result, which could be caused by the severe noise in dFC time series. In contrast, our method could learn the intrinsic brain status transition, thus is more robust to such noise. 

To validate the advantage of Full-BiLSTM, we tested three other LSTM-based architectures. The BiLSTM\_Last model uses the output of the last BiLSTM cell for classification, as used in the traditional sequence processing studies. The Full-LSTM uses the same architecture as our method, but with uni-directional LSTM cells. To investigate whether a deeper BiLSTM layer could increase the performance, the third model is built using stacked Full-BiLSTM (two layers). All these three models use the same parameters as our Full-BiLSTM method. As shown in Fig.~\ref{Fig3})., our model still outperformed all these three LSTM-based competing models. Specifically, we observed that 1)BiLSTM outperforms uni-directional LSTM; 2)Full-BiLSTM performs better than BiLSTM\_Last; 3)A deeper model does not improve the final performance. In addition, we also compared the performance with and without data augmentation, and found that the accuracy was decreased by 2\% without data augmentation. Furthermore, the number of hidden nodes in LSTM may directly affect the learning capacity of an LSTM network. Therefore, we compared the performance of Full-BiLSTM models with a varying number of hidden units, i.e., {16, 32, 64}. As shown in Fig.~\ref{Fig4}, the Full-BiLSTM model with 16 hidden nodes has decreased performance and increased performance variability, compared to the Full-BiLSTM model with 32 hidden nodes. It is likely that 16 hidden units are too limited to store the sequential information of the dFC process. The model with 64 hidden nodes also has suboptimal performance, which could be attributed to overfitting.

The results together indicate that data augmentation and the choice of network structure are crucial for training an effective dFC-based classification model. Most notably, this is the first attempt to use a deep learning framework for individualized disease diagnosis based on dFC. Our results indicate that a sequence model can take advantage of more series information from dFC than the conventional methods. It is also worth noting that our model can be applied to other brain disorder diagnoses.

\vspace{-0.2cm}  

\section{Conclusions}

\vspace{-0.2cm}
In this study, we proposed a new deep learning framework, a Full-BiLSTM model, for brain disease diagnosis using dynamic functional connectivity. To the best of our knowledge, this is the first attempt to propose the “deep chronnetome learning” framework and to prove its feasibility and superiority in a challenging MCI diagnosis task by using time-varying functional information. Comprehensive comparisons among different architectures of the LSTM model were conducted, and the insightful discussions on the influence of the hyperparameters were provided. In summary, the proposed model can not only effectively capture the trait-related brain dynamic changes from the spatiotemporally complex chronnectome, but also can be applied to improve classification of other brain disorders, which shows great promise to be used as a powerful tool to detect potential biomarkers in the community.

\end{document}